\documentclass{article}

\usepackage[final]{neurips_2019}

\usepackage[utf8]{inputenc} 
\usepackage[T1]{fontenc}    
\usepackage{hyperref}       
\usepackage{url}            
\usepackage{booktabs}       
\usepackage{amsfonts}       
\usepackage{nicefrac}       
\usepackage{microtype}      

\setcitestyle{square}

\title{A Step Towards Exposing Bias in Trained Deep Convolutional Neural Network Models}
\usepackage{graphicx}
\usepackage{subcaption}
\author{%
  Daniel Omeiza \\
  \\
  \texttt{daniomeiza@gmail.com}
}

\begin{document}

\maketitle

\begin{abstract}
 Machine learning algorithms could be bias on classes like race and gender. For example, face recognition model trained on non-inclusive dataset can cause a chain of errors, leading to the mis-identification of perpetrators from security video footage analysis in the criminal justice system. Hence, gaining insight into how Deep Convolutional Neural Network(CNN) models perform image classification and how to explain their outputs is a concern. As an effort to developing explainable deep learning models, several methods have been proposed such as finding gradients of class output with respect to input image (sensitivity maps), class activation map (CAM), and Gradient based Class Activation Maps (Grad-CAM). These methods under perform when localizing multiple occurrences of the same class and do not work for all CNNs. In addition, Grad-CAM does not capture the entire object in completeness when used on single object images. We present Smooth Grad-CAM++, a technique which combines two recent techniques: SMOOTHGRAD and Grad-CAM++. Smooth Grad-CAM++ has the capability of either visualizing a layer, subset of feature maps, or subset of neurons within a feature map at each instance. We experimented with few images, and we discovered that Smooth Grad-CAM++ produced more visually sharp maps with larger number of salient pixels highlighted in the given input images when compared with other methods. Smooth Grad-CAM++ will give insight into what our deep CNN models (including models trained on medical scan or imagery) learn.  Hence informing decisions on creating a representative training set.
\end{abstract}

\section{Introduction}
AI-based technologies that are not specifically trained to perform high stake tasks, however, they can be used in processes where such tasks are involved. It is very likely that face recognition software by itself is being used to identify suspects, although such software should not be trained to determine the fate of individual in the criminal justice system. This causes chains of error from the erroneously confident software, leading to the mis-identification of perpetrators from security video footage analysis in the criminal justice system \citep{b1}.
As much as the performance of some complex models in computer vision are improved, they may posses limitations from their inability to explain their decisions to human users. This is a non-trivial issue in risk-averse domains such as in security, health and autonomous navigation \citep{bb1}. 
Visualization techniques such as saliency (sensitivity or pixel attribution) \citep{bb1,b12,b13}, general Class Activation Map (CAM) \citep{bb1} under perform when localizing multiple occurrences of the same class. In addition, Gradient based CAM (Grad-CAM) \citep{b32} does not capture an entire object in completeness when used on single object images. These affect performance on recognition tasks. Although, Grad-CAM++ technique tends to take care of these limitations, improvements are required in terms of class object capturing and visual appeal. Furthermore, the implementations of the current visualization techniques do not go down to the neuron level (we may be sometimes interested in a group of neurons).

In this paper, we introduce gradient smoothening into Grad-CAM++, the resulting technique makes provision for visualizing a convolution layer, subset of feature maps and subset of neurons in a feature map with improved visual appeal and class object capturing.
Smoothening entails adding Gaussian noise to the sample image of interest, and  taking the average of all gradient matrices generated from each noised image.

\section{Background}
\subsection{Algorithm Bias}
The evaluation of bias in automated facial analysis algorithms and datasets with respect to phenotypic subgroups shows an uneven distribution of skin colors in dataset. The datasets are overwhelmingly composed of lighter-skinned subjects with 79.6\% for IJB-A and 86.2\% for Adience \citep{b1}. Upon the evaluation of 3 commercial gender classification systems using a corrected and well representative dataset, results showed that darker-skinned females are the most misclassified group with error rates
of up to 34.7\%. The maximum error rate
for lighter-skinned males was 0.8\% \citep{b1}. This calls for urgent attention if we care about fair AI systems.

\subsection{Efforts Towards Explainability}
Explainability is believed to expose the flaws in computer vision powered systems. Some earlier proposed techniques include: Local Interpretable Model-Agnostic Explanations(LIME) \citep{b18}, DeepLift \citep{b19}, Contextual Explanation Networks (CENs) \citep{b20}. 
More recent visualization techniques in the order of improved performance are Class Activation Map (CAM), Gradient-Weighted Class Activation Map(Grad-CAM) and a generalization of Grad-CAM called Grad-CAM++ \citep{b31}.

While CAM is limited to a narrow class of CNN models, Grad-CAM is broadly applicable to any CNN-based architectures and needs no re-training.

Grad-CAM++ uses the weighted combination of the positive partial derivatives of the last convolution layer feature maps with respect to a specific class score as weights to generate a visual explanation for the class label under consideration. The author formulated 
    \(W^{c}_{k}\) to capture the importance of a particular activation map \(A_{k}\) by:
    \begin{equation}
        W^{c}_{k} = \sum_{i}\sum_{j}\alpha^{kc}_{i,j} ReLU\left(\frac{\partial Y^{c}}{\partial A^{k}_{i,j}}\right)
        \label{eq:activationmap}
    \end{equation}
where \( \alpha^{kc}_{i,j} \) captures the importance of location \((i,j)\) for activation map \(A^{k}\) for target class \(c\).
The heatmap is a weighted combination of feature maps with ReLU.
    \begin{equation}
        L^{c}_{Grad-CAM} = ReLU\left(\sum_{k}{W^{c}_{k} A^{k}}\right)
        \label{eq:ReLU}
    \end{equation}
    
\citep{b8} introduced SMOOTHGRAD, a simple method that can help visually sharpen gradient-based sensitivity maps by taking random samples in a neighborhood of an input x, and averaging the resulting sensitivity maps. Mathematically, this
means calculating:
\begin{equation}
    M_c\left(x\right) = \frac{1}{n}\sum_{1}^{n}M_{c}\left(x + \mathcal{N}\left(0, \sigma^2\right)\right)
    \label{eq:smoothgrad}
\end{equation}
where \(n\) is the number of samples, and \( \mathcal{N}\left(0, \sigma^2\right) \) represents Gaussian noise with standard deviation \(\sigma\).
With the motivation to provide an enhanced visualization maps, we apply this smoothening technique in the gradients computations involved in Grad-CAM++ as shown above. The resulting gradients are applied in the Grad-CAM++ algorithm. This provides better (in terms of visual appeal and capturing) maps for deep CNNs.

\section{Methodology}
\subsection{Noise Over Input }
\(n\) noised samples are generated by adding Gaussian noise to original using following the standard mean deviation value \(\sigma\) (the degree of noise to be added). \(\sigma\) could be varied till a satisfactory visual map is produced.
\subsection{Gradients Averaging}
The averages of all \(1st\), \(2nd\) and \(3rd\) order partial derivatives of all \(n\) noised inputs are taken and are used in computing \(\alpha^{kc}_{i,j}\) and \( W^{c}_{k}\).

Let \(D_{1}^{k}, D_{2}^{k} \) and \(D_{3}^{k} \) denote matrices of \(1st\), \(2nd\) and \(3rd\) order partial derivatives respectively for feature map \(k\).
With reference to the original Grad-CAM++ formulation, we can now compute \(\alpha^{kc}\) as:
\begin{equation}
    \alpha^{kc}_{i,j} = \frac{\frac{1}{n}\sum_{1}^{n}D_{1}^{k}}{2\frac{1}{n}\sum_{1}^{n}D_{2}^{k} + \sum_{a}\sum_{b} A^{k}_{a,b} \frac{1}{n}\sum_{1}^{n}D_{3}^{k}}
    \label{eq:computed_a}
\end{equation}
and substituting the averaged gradient into Equation \ref{eq:activationmap}, Grad-CAM++ weights \( W^{c}_{k}\) becomes:
\begin{equation}
    W^{c}_{k} = \sum_{i}\sum_{j}\alpha^{kc}_{i,j} ReLU\left(\frac{1}{n}\sum_{1}^{n}D_{1}^{k}\right)
    \label{eq:gradplus}
\end{equation}
When \(W^{c}_{k}\) is substituted into Equation \ref{eq:ReLU}, we get the final class discriminative saliency matrix which could be plotted with matplotlib or any other image plot library. This serves as the final saliency map. This modified Grad-CAM++ is what we call Smooth Grad-CAM++. We provide an Application Programming Interface (API) for this.

\subsection{Choosing Parameter Values}
At each instance, only one convolution layer can be visualized. The name of the layer to be visualized is passed to the API. Names by default have a specific convention, however, viewing the summary of the trained model will reveal the name or unique identifier of each convolution layer.

To specify the feature maps to be visualized, the filter parameter must be set. The filter parameter is a list of integers specifying the index of the feature maps to be visualized in the specified convolution layer. If \(n\) values are set for the filter parameter, \(n\) maps are generated corresponding to each feature map.

Visualizing neurons is useful when individual neuron activation is of interest. For instance, \citep{b21} used subset scan algorithm to identify anomalous activations in a convolution neural network. Smooth Grad-CAM++ will be useful in providing good explanations at the neuron level.
Our API provides an option to visualize regions of neurons within a specified coordinate boundary when region parameter is set to true. When region parameter is set to false and a subset of coordinates is provided, only the neurons in those coordinates are visualized while other activations are clipped at zero.

Any learned deep CNN model can be chosen for visualization. In this paper, we used VGG-16 pre-trained model and explored the last convolution layer.

\begin{figure}[htbp]
  \centering
  \begin{subfigure}[b]{0.5\linewidth}
    \includegraphics[width=\linewidth]{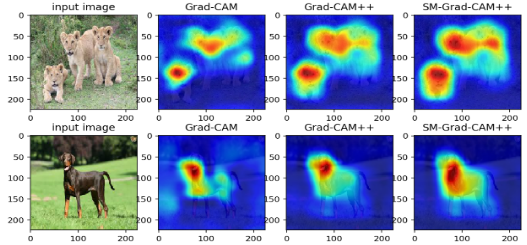}\hfill
    \caption{Last Conv layer visualization with different techniques}
  \end{subfigure} 
\end{figure}

\begin{figure}[htbp]
  \centering
  \begin{subfigure}[b]{0.4\linewidth}
    \includegraphics[width=\linewidth]{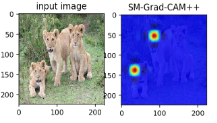}\hfill
    \caption{Neurons at coordinates (3,5) and (5,5) of feature map 10.}
  \end{subfigure}
   \begin{subfigure}[b]{0.4\linewidth}
    \includegraphics[width=\linewidth]{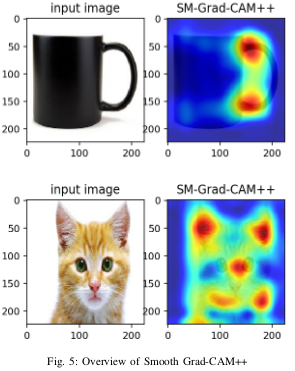}\hfill
    \caption{Saliency detection at inference stage}
  \end{subfigure}
  
  \caption{Saliency maps generated using Smooth Grad-CAM++}
  \label{fig:node1}
\end{figure}

\section{Results}
From Figure 1, Smooth Grad-CAM++ gives a clearer explanation of particular features the model learned. 
Figure 1 shows comparison with different visualization algorithms and our proposed algorithm (Smooth Grad-CAM++). Figure 2a shows the output of a neuron visualization with Smooth Grad-CAM++ using VGG-16 trained pre-trained model. Figure 2a shows the output for the input image of a coffee mug and a cat with the same VGG-16 pre-trained model. From visual inspection of the generated maps, we see justifications for the model's predictions. The map highlighted salient features in the input image which it has previously learned during training to inform its prediction process.

\section{Conclusion}
Enhanced visual saliency maps can help demystify the internal workings of deep convolution neural network models and give computer vision researchers better insights. In this paper, we proposed Smooth Grad-CAM++, an enhanced visual map for deep convolution neural networks. Our experiment results disclosed an improvement in the generated maps when compared with existing methods. These maps were generated by averaging gradients (i.e derivative of class score with respect to the input) from many small perturbations of a given image and applying the resulting gradients in the generalized Grad-CAM algorithm (Grad-CAM++). Smooth Grad-CAM++ performs well in object localization and also in multiple occurrences of an object of same class. It is able to create maps for specific layers, subset of feature maps and neurons of interest, and it is able to highlight larger number of salient pixels in the maps.  Our technique (Smooth Grad-CAM++ with API provided) justifies the decisions of deep CNN models. Therefore, it is easier to tell when a model has been trained on a non-inclusive dataset by testing with the presumed unrepresented object. This concept of explainability basically help to create awareness on the existence of biases in deep CNN models, and also informs against the deployment of biased models. Smooth Grad-CAM++ can help expose bias in models trained on medical scans or imagery. For instance, employing it to confirm if a specific class of data was represented in the training set for a learned model. This is  especially important for a model trained on a skin disease dataset where skin color matters. The limitation of this work is that at present, it only applies to deep CNN models. Also, there exist no quantitative means for evaluating performance. The risk in this work is that there could be some level of uncertainty concerning the existence of bias in a trained model since there is no quantitative way to measure its presence. As potential solutions to eradicating bias in a biased model is not considered, a potential effect is that people could keep on using biased models because they lack ideas on how to remove the biases. Future works involve generalizing Smooth Grad-CAM++ to work for other network architectures other than deep CNNs, and devising a quantitative means of measuring the effectiveness of generated maps in pointing out bias. We shall also add suggestions on how to remove bias when they are discovered.

\end{document}